\documentclass[a4paper,fleqn]{cas-dc}

\usepackage[authoryear]{natbib}
\usepackage{amsmath}
\DeclareMathOperator*{\argmax}{arg\,max}
\def\tsc#1{\csdef{#1}{\textsc{\lowercase{#1}}\xspace}}
\tsc{WGM}
\tsc{QE}

\begin{document}
\let\WriteBookmarks\relax
\def\floatpagepagefraction{1}
\def\textpagefraction{.001}

\shorttitle{}    

\shortauthors{}  

\title [mode = title]{Style-Decoupled Adaptive Routing Network for Underwater Image Enhancement}  



%

\author[1]{Hang Xu}[orcid=0009-0004-0577-0116]
\ead{190107xh@whu.edu.cn}
\credit{Writing - Original Draft, Methodology, Visualization}

\author[1]{Chen Long}
\cormark[1]
\ead{chenlong107@whu.edu.cn}
\credit{Writing - Review \& Editing, Conceptualization}

\author[2]{Bing Wang}
\ead{bingwang@polyu.edu.hk}
\cormark[1]
\credit{Writing - Review \& Editing, Supervision}

\author[1]{Hao Chen}
\ead{silverchen2024@whu.edu.cn}
\credit{Writing - Review \& Editing}

\author[1]{Zhen Dong}
\ead{dongzhenwhu@whu.edu.cn}
\credit{Writing - Review \& Editing, Supervision, Funding Acquisition, Resources}


\affiliation[1]{organization={State Key Laboratory of Information Engineering in Surveying, Mapping and Remote Sensing, Wuhan University},
            city={Wuhan},
            postcode={430079}, 
            country={China}}

\affiliation[2]{organization={Department of Aeronautical and Aviation Engineering, The Hong Kong Polytechnic University},
            city={Hung Hom},
            state={Kowloon},
            country={Hong Kong Special Administrative Region of China}}

\cortext[1]{Corresponding author.}



\begin{abstract}
Underwater Image Enhancement (UIE) is essential for robust visual perception in marine applications. However, existing methods predominantly rely on uniform mapping tailored to average dataset distributions, leading to over-processing mildly degraded images or insufficient recovery for severe ones.
To address this challenge, we propose a novel adaptive enhancement framework, SDAR-Net. Unlike existing uniform paradigms, it first decouples specific degradation styles from the input and subsequently modulates the enhancement process adaptively. 
Specifically, since underwater degradation primarily shifts the appearance while keeping the scene structure, SDAR-Net formulates image features into dynamic degradation style embeddings and static scene structural representations through a carefully designed training framework.
Subsequently, we introduce an adaptive routing mechanism. By evaluating style features and adaptively predicting soft weights at different enhancement states, it guides the weighted fusion of the corresponding image representations, accurately satisfying the adaptive restoration demands of each image. Extensive experiments show that SDAR-Net achieves a new state-of-the-art (SOTA) performance with a PSNR of 25.72 dB on real-world benchmark, and demonstrates its utility in downstream vision tasks. Our code is available at \url{https://github.com/WHU-USI3DV/SDAR-Net}.

\end{abstract}





\maketitle

\section{Introduction}\label{Introduction}
Underwater imaging \citep{li2025underwater} plays a crucial role in various applications, including marine exploration \citep{mittal2022survey}, autonomous underwater vehicle navigation \citep{leonard2016autonomous,zhang2025spatial,wang2024codeunet}, ecological monitoring \citep{shortis2016review}, and underwater archaeology \citep{singh2000imaging}. However, underwater images are inherently degraded due to wavelength-dependent absorption and particle scattering, which typically manifest as severe color distortion, diminished contrast, and structural blurring \citep{liang2021single,song2024advanced}. Such degradation phenomena can adversely affect the performance of downstream vision tasks, such as object detection \citep{lei2022underwater,wittmann2024robust}, semantic segmentation \citep{islam2020semantic,qin2025causal}, and 3D reconstruction \citep{hu2023overview,zhong2025high,ma2025cross,zhong2025cutting}. Consequently, Underwater Image Enhancement (UIE) has become a pivotal research topic that bridges computer vision and marine engineering, serving both as a preprocessing module for automated systems and as a standalone solution for visual quality restoration \citep{raveendran2021underwater}.

\begin{figure*}
	\centering
	\includegraphics[width=1\textwidth]{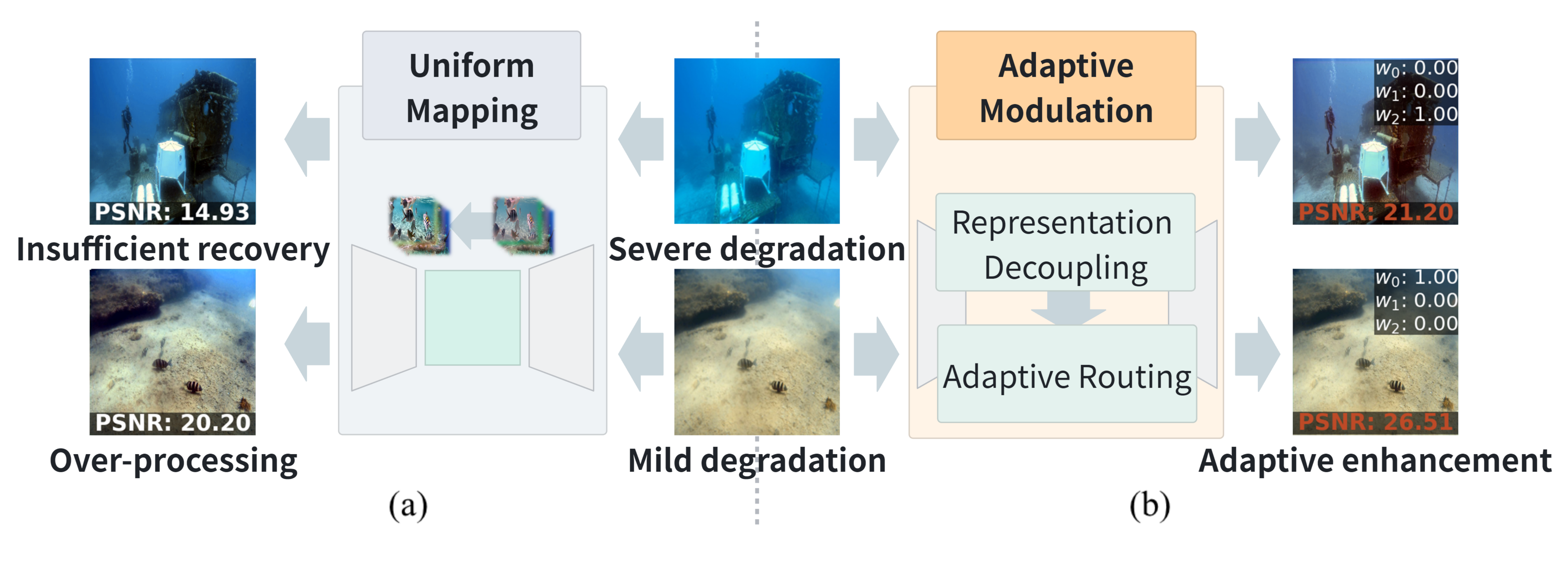}
	\caption{
    (a) Previous methods applying uniform enhancement strategies yield suboptimal results across various degradation degree. (b) Our approach adaptively modulates the enhancement process under the guidance of degradation patterns, addressing the edge-cases and enhancing overall performance.
    }
	\label{FIG:1}
\end{figure*}

Recent advances in UIE field \citep{HCLR-Net, U-shape, DW-net, GUPDM, Semi-UIR} have achieved remarkable progress by exploring various deep learning architectures.
Despite their average effectiveness, the dominant paradigm remains rooted in learning a deterministic mapping that approximates the average degradation distribution of the training data. This uniform approach inherently lacks the adaptability to handle the diverse underwater image degradations. By predominantly minimizing a global loss over a diverse training set, these models converge toward a uniform solution that is sub-optimal for individual instances. As illustrated in Fig. \ref{FIG:1} (a): a uniformed mapping tends to over-process mildly degraded images, 
while providing insufficient restoration for severely degraded images.

To address these limitations, we propose an adaptive UIE architecture named Style-Decoupled Adaptive Routing Network (SDAR-Net). As shown in Fig. \ref{FIG:1} (b), this architecture decouples varying degradation patterns from invariant scene structures in the enhancement process, enabling adaptive modulation of its enhancement behavior. 

The first step of SDAR-Net is to model the varying degradation patterns of underwater images. 
Specifically, motivated by the observation that underwater degradation manifests as environment-induced appearance variations while the scene content remains relatively stable \citep{Ucolor, islam2020fast}, we abstract the underwater imaging process as $\mathcal{C} = \text{F}(\mathcal{S}, \mathcal{B})$. Here, $\mathcal{C}$ denotes the image representation, $\mathcal{S}$ denotes the varying degradation patterns, $\mathcal{B}$ represents the invariant scene structure, while $\text{F}$ is the imaging process. Consequently, the primary objective of the first step is to thoroughly disentangle $\mathcal{S}$ from the image representation $\mathcal{C}$, which is realized through a style-transfer-inspired training paradigm. 


With the degradation patterns explicitly modeled, we proceed to modulate the enhancement process via an adaptive routing mechanism. Motivated by the observation that diverse underwater scenes require varying restoration intensities, we conceptualize UIE not as a uniform mapping, but as a continuous state evolution driven by a recursive refinement process. To realize this evolution, we utilize the enhancement backbone as a shared recursive unit to generate multiple candidate states that represent a sequence of potential restoration outcomes. 
By evaluating the decoupled style representations of these candidates, the router predicts soft-weighted probabilities to adaptively synthesize a unique restoration behavior for each input. This approach enables SDAR-Net to transcend the average training distribution, effectively handling edge cases and extreme degradations with significantly improved robustness and content fidelity.

The main contributions of this work are listed as follows:

(1) We analyze the limitations of uniform mapping in existing UIE methods and propose an adaptive framework to address these problems, including 
representation decoupling and adaptive trajectory modulation.


(2) We implement this dynamic paradigm through SDAR-Net, which integrates two core components: A representation decoupling framework that explicitly models degradation patterns and an adaptive routing mechanism that dynamically synthesizes the optimal enhancement behavior.

(3) Extensive experiments on multiple public real-world benchmarks demonstrate that the proposed method consistently outperforms current state-of-the-art methods in both qualitative and quantitative accuracy, with significant gains in downstream tasks.

\section{Related works}\label{Related Works}
The goal of UIE is to restore visually plausible and perceptually natural images by compensating for degradations caused by wavelength-dependent light absorption and scattering in underwater environments. Over the past two decades, UIE approaches have transitioned from physics-inspired analytical models\citep{pizer1987adaptive,pizer1990contrast,he2010single,drews2013transmission,drews2016underwater,GUDCP,ancuti2012enhancing,wang2023meta,wang2024self} to data-driven training paradigms. In this section, we introduce the current learning-based UIE methods organized as single-stage methods (Sect.~\ref{subsec:rws}), multi-stage methods (Sect.~\ref{subsec:rwmoe}).


\subsection{Single-stage methods}\label{subsec:rws}
Single-stage methods aim to establish a complete mapping from degraded inputs to enhanced outputs through a unified network architecture, without explicit functional modularization or multi-stage expert decomposition. The advent of deep learning has significantly advanced UIE by enabling the end-to-end learning of complex mappings. Early pioneering efforts established foundational datasets \citep{UIEB,U-shape,UWCNN} to facilitate supervised training. Since then, a vast body of work has focused on direct regression using diverse architectural innovations, ranging from CNNs \citep{krizhevsky2012imagenet} and U-net \citep{ronneberger2015u} structures \citep{PRWNet, SCNet, SGUIE, UICoE, UIEWD, PUIE, LCNet,URanker,SFGnet,UIR-PK,DW-net,HCLR-Net,ye2026jdpnet} to Transformers \citep{vaswani2017attention} and Mamba \citep{gu2024mamba} frameworks \citep{URSCT,X-CAUNET,SS-UIE,uwmamba}.
Furthermore, to address the scarcity of paired data, unsupervised and semi-supervised frameworks like USUIR \citep{USUIR} and Semi-UIR \citep{Semi-UIR} have been proposed to exploit unpaired data through cycle consistency and self-supervision.

Generative frameworks have been widely adopted to improve perceptual quality and generalization. Generative Adversarial Networks (GANs) based methods \citep{UGAN,watergan,U-shape,FUnIE,CWR,UIE-DAL,P2Cnet,CLUIE,PUGAN,TUDA}
are designed to produce more natural-looking results through adversarial training. More recently, diffusion-based methods \citep{WF-Diff,SU-DDPM,DM-underwater,Patch-UIE-DIFF,uiedp} 
have emerged as powerful alternatives, offering superior sample fidelity by iteratively denoising from learned data distributions. In recent years, video diffusion models have also been used to address consistent enhancement of serialized underwater images \citep{hu2025underwater}. To further support these generative models, data synthesis strategies like Osmosis \citep{osmosis} and SLURPP \citep{SLLRUP} leverage physics-based simulators to generate realistic training pairs.

While these methods maintain a single-stage pipeline, some incorporate internal strategies such as transmission or frequency decomposition to better exploit intrinsic image properties. For instance, GUPDM \citep{GUPDM} and DPF-Net \citep{mei2025dpf} explicitly models depth-dependent attenuation to regularize the restoration process. WF-Diff \citep{WF-Diff} uses frequency decomposition to obtain the conditions for guided diffusion. These methods often rely on fixed prior patterns, which keep their architecture within the scope of a uniform mapping.

Despite these advances, existing single-stage methods often rely on a uniform enhancement mapping that approximates the average data distribution. While computationally efficient and straightforward, this approach struggles to fully exploit the dataset's diversity, failing to handle edge cases and extreme degradations.

\subsection{Multi-stage methods}\label{subsec:rwmoe}
Given the inherent complexity and diverse degradation factors of underwater environments, several works have moved beyond single-stage architectures toward multi-stage frameworks. A typical multi-stage decomposition includes the decomposition of the UIE task workflow or the decomposition of UIE task types. Some multi-stage methods typically decompose the restoration task into sequential, functionally distinct modules. For instance, CA-Net \citep{CAnet} adopts a coarse-to-fine strategy by performing feature-level restoration followed by a dedicated color correction network. Similarly, CCL-Net \citep{CCLnet} first utilizes color decomposition for chromatic correction and subsequently addresses residual haziness through a specialized dehazing module.

In parallel, other multi-stage architectures aim to handle diverse degradations by processing inputs through specialized network streams. This is often achieved through feature-driven branching, which examines the input from various physical or color-space perspectives. A representative work, WaterNet \citep{UIEB}, feeds the network with parallel branches for white balance, histogram equalization, and gamma correction. Similarly, Ucolor \citep{Ucolor} and UIEC$^2$-Net \citep{UIEC2-net} leverage multi-stream encoders to explicitly process images in RGB, Lab, and HSV color spaces, aiming to decouple chromatic restoration from structural enhancement. 
PRWnet \citep{PRWNet} and UIEWD \citep{UIEWD} utilize wavelet decomposition to synchronously process distortions in high-frequency texture and low-frequency color information. Alternatively, degradation-specific branching assigns distinct modules to handle different types of degradation. For example, UWCNN \citep{UWCNN} trains ten separate models tailored to specific Jerlov water types \citep{jerlov1968optical}; however, these models operate independently and lack a unified architecture to adaptively combine their strengths during inference. More recently, UniUIR \citep{UniUIR} addresses this by incorporating a Mixture-of-Experts (MoE) \citep{jacobs1991adaptive,pioro2024moe} architecture, which adaptively activates learnable expert branches within a single framework based on the input's characteristics.

Existing multi-stage approaches focus on improving the interpretability and perceptual quality by decomposing the UIE task to subtasks such as color correction, dehazing, and water-type-based divisions. In contrast, our method introduces an adaptive architecture to address edge-cases in diverse underwater environments. This design broadens the methodology to include instance-aware precision with explicit degradation modeling and adaptive enhancement process modulation.

\begin{figure*}
	\centering
	\includegraphics[width=1.0\textwidth]{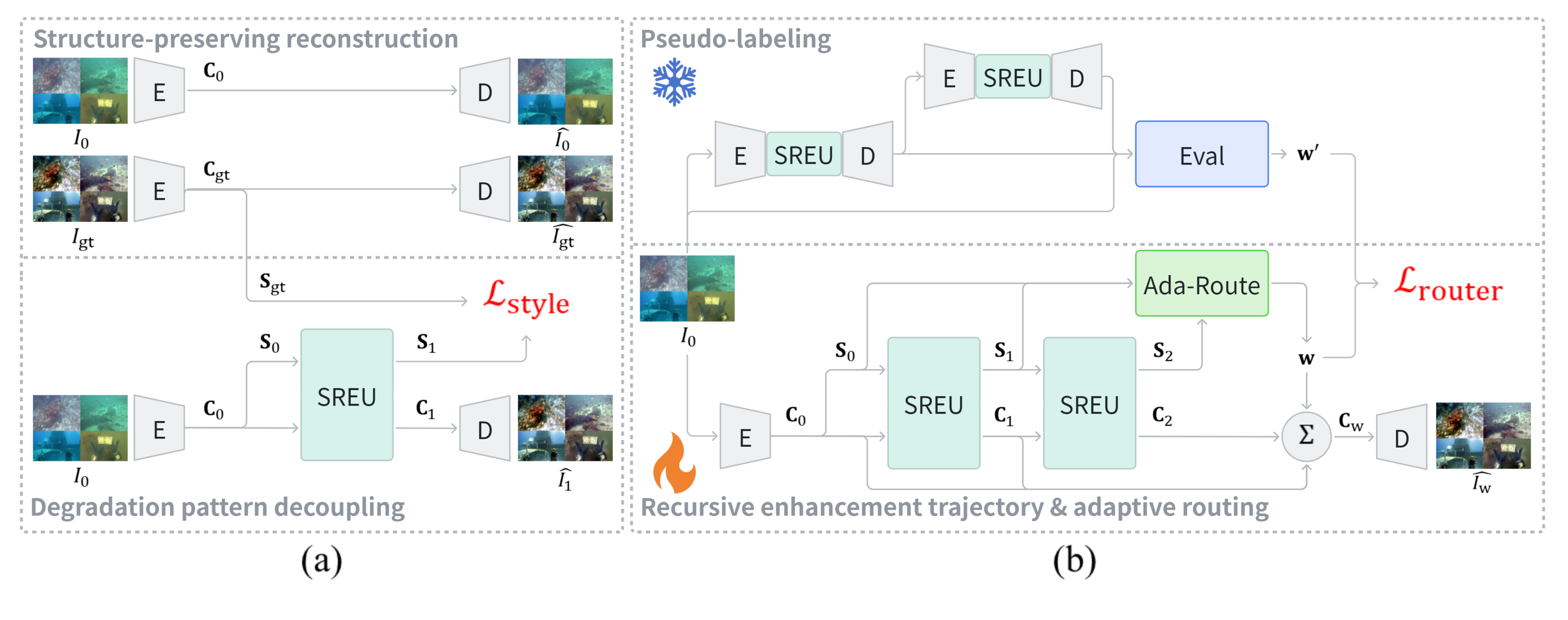}
	\caption{Overall framework of SDAR-Net. This framework consists of two parts: (a) Representation Decoupling; (b) Adaptive Trajectory Modulation.}
	\label{FIG:2}
\end{figure*}

\section{Methodology}\label{Methodology}
In this section, we detail the proposed SDAR-Net framework for adaptive underwater image enhancement. To overcome the suboptimal enhancement problem of conventional uniform mappings, our approach formulates restoration as an adaptive modulation task. 
As illustrated in Fig. \ref{FIG:2}, the overall architecture comprises two primary components: (1) A representation decoupling process, which decouples degradation patterns from image representations through a training framework that includes multiple training objectives (Sect.~\ref{subsec:mst}); (2) An adaptive trajectory modulation mechanism that adaptively adjusts the enhancement trajectory according to the decoupled degradation pattern (Sect.~\ref{subsec:mar}). 

\subsection{Representation decoupling}\label{subsec:mst}

To develop an enhancement architecture that adaptively responds to image degradation, we first decouple degradation patterns from the image and model their evolution process. We observed that underwater degradation primarily affects appearance attributes such as color distribution and contrast ratio, while the scene structure remains invariant. 
Thus, we characterize the degradation pattern through style features that capture the image's appearance attributes. 

Specifically, we first obtain robust image representations that anchored to scene structure through reconstruction tasks with a lightweight encoder-decoder (E and D). 
Then we obtain degradation-aware style features from the image representations and model the process of transferring degraded style features to clear style features through a shared recursive enhancement unit (SREU). In this process, the style features synchronously guide the image representations to perform the transformation from degraded to clear. 
To ensure that style features can actually characterize the degradation patterns, we draw inspiration from image style transfer and design a transfer loss to guide the modeling of degradations-related style. 


\subsubsection{Structure-preserving reconstruction}\label{subsec:msr}
\begin{figure}
	\centering
	\includegraphics[width=1.0\columnwidth]{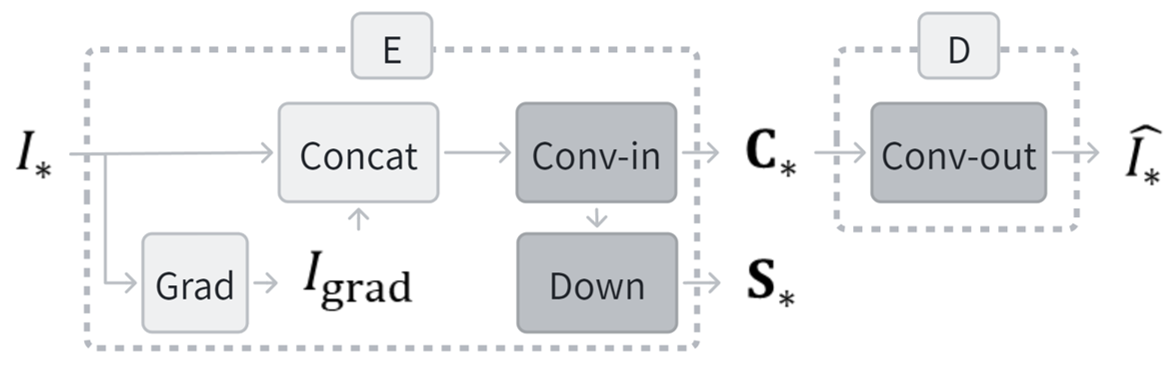}
	\caption{Design of lightweight encoder (E) and decoder (D). Conv-in and Conv-out are convolutional layers, Grad denotes the gradient extraction operation and Down denotes downsample operation. }
	\label{FIG:3}
\end{figure}

To obtain robust image representations that can accurately restore scene structure, we train an image reconstruction task with a lightweight encoder-decoder structure. The design of the encoder (E) and decoder (D) is shown in Fig \ref{FIG:3}, where \( {I}_*, \hat{I_*} \) denotes the input and reconstructed image and \( \textbf{C}_* \) denotes the feature for image representation $\mathcal{C}$. 
In order to anchor the image representation to the scene structure in design, we incorporate gradient-based information during the encoding process, since the invariant scene structure is highly correlated with edge features. The Grad operation converts the image to grayscale and extracts the gradient \( {I}_\text{grad} \) of the image  through a gradient convolution. To ensure structural integrity, we use simple convolutional layers to extract $\textbf{C}_*$ and decode $\hat{I_*}$.

To enforce structural fidelity during reconstruction, we define a multi-component reconstruction loss between two images \( {I}_* \) and \( \hat{I_*} \). First, the basic pixel-wise loss is defined as: 
\begin{equation}
\mathcal{L}_{\text{str}}({I}_*, \hat{I_*}) = \lambda_{\text{L1}} \mathcal{L}_{\text{1}}({I}_*, \hat{I_*}) + \lambda_{\text{L2}} \mathcal{L}_{\text{2}}({I}_*, \hat{I_*}),
\end{equation}
where \(\mathcal{L}_{\text{1}} \), \(\mathcal{L}_{\text{2}} \) denotes the L1 loss and the MSE loss, and $\lambda_{\text{L1}}$, $\lambda_{\text{L2}}$ denotes the weight assigned to each loss.

Second, to further emphasize structural consistency, the gradient-aware loss is defined as:
\begin{equation}
\mathcal{L}_{\text{grad}}({I}_*, \hat{I_*}) = \mathcal{L}_{\text{1}}(\text{Grad}({I}_*), \text{Grad}(\hat{I_*})) .
\end{equation}

Third, to make the reconstruction more natural and fidelity, we incorporate the following perceptual loss following \citep{Semi-UIR,U-shape}:
\begin{equation}
\mathcal{L}_{\text{perc}}({I}_*, \hat{I_*}) =  ||\phi({I}_*) - \phi(\hat{I_*})||_2,
\end{equation}
where \( \phi(\cdot) \) denotes the feature map extracted from a pre-trained VGG-19 \citep{simonyan2014very} network.

Combining all the aforementioned losses, the final reconstruction loss is formulated as follows and serves as the foundation for all subsequent image-related loss terms.
\begin{equation}
\begin{split}
\mathcal{L}_{\text{recon}}({I}_*, \hat{I_*}) & = \lambda_{\text{str}} \mathcal{L}_{\text{str}}({I}_*, \hat{I_*}) + \lambda_{\text{grad}} \mathcal{L}_{\text{grad}}({I}_*, \hat{I_*}) \\
& + \lambda_{\text{perc}} \mathcal{L}_{\text{perc}}({I}_*, \hat{I_*}).
\end{split}
\end{equation}
where $\lambda_{\text{str}}$, $\lambda_{\text{grad}}$ and $\lambda_{\text{perc}}$ denote the weights assigned to each loss. 

We used both the input images and the ground truth images of the dataset to train the reconstruction process. Specifically, we execute the processes of $\textbf{C}_0,\textbf{S}_0=\text{E}(I_0)$ and $\textbf{C}_\text{gt},\textbf{S}_\text{gt}=\text{E}(I_\text{gt})$, then decode the reconstructed images as $\hat{I_0}=\text{D}(\textbf{C}_0)$ and $\hat{I_\text{gt}}=\text{D}(\textbf{C}_\text{gt})$. Finally, we calculate both $\mathcal{L}_{\text{recon}}({I}_0, \hat{I_0})$ and $\mathcal{L}_{\text{recon}}({I}_\text{gt}, \hat{I_\text{gt}})$.

\subsubsection{Degradation pattern decoupling}\label{subsec:mstg}
Following the extraction of robust image representations, we continue to acquire the degradation-aware style feature $\mathbf{S}_*$ to characterize the degradation pattern $\mathcal{S}$. As illustrated in Fig. \ref{FIG:3}, $\mathbf{S}_*$ is simply obtained via a downsample operation (Down) comprising average pooling and convolutional layers. 
To ensure that $\textbf{S}_*$ can actually characterize the decoupled degradation pattern and decoupled from scene structure, we proposed a dual strategy combining architecture constraints and loss constraints. 

\begin{figure}
	\centering
	\includegraphics[width=0.95\columnwidth]{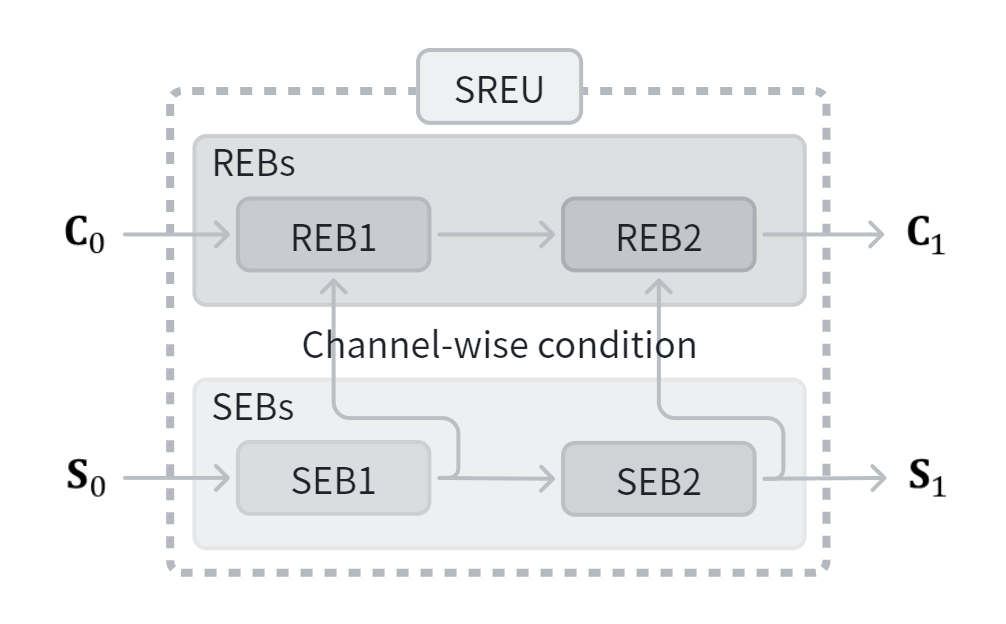}
	\caption{Design of SREU. REB denotes Representation Enhancement Block, SEB denotes Style Evolution Block.}
	\label{FIG:4}
\end{figure}

In terms of architecture constraints, since the encoder has already anchored the scene structure within the image representation, we employ a decoupled architecture in SREU to separately execute the evolution of $\textbf{S}_0 \rightarrow \textbf{S}_1$ and the enhancement process of $\textbf{C}_0 \rightarrow \textbf{C}_1$. 
As illustrated in Fig. \ref{FIG:4}, we reformulate the representation enhancement process by treating the evolving style features as a conditional prior. Inspired by the efficiency of module designs in Semi-UIR \citep{Semi-UIR}, we adopt its basic block structures for our Representation Enhancement Blocks (REBs) and Style Evolution Blocks (SEBs). Based on this, our core innovation lies in the architectural separation and the residual condition injection. 
Specifically, style features extracted by the SEBs serve as a conditional prior for the REBs. To ensure structural decoupling, these features are upsampled via bilinear interpolation and integrated through $1 \times 1$ convolutions followed by channel-wise residuals. This mechanism allows the style condition to guide the image representation enhancement without influencing the scene structure.
This design architecturally isolates the scene-structure-anchored representations from the degradation-aware style features, enabling decoupled modeling of the evolution process of the degradation pattern.


For loss constraints, 
we supervise the evolution process by aligning the channel-wise similarity between the evolved style feature $\textbf{S}_1$ and the ground-truth style feature $\textbf{S}_\text{gt}$, thus characterizing the transformation of degradation patterns. We introduce a style decoupling loss based on the Gram matrix \citep{gatys2015texture}, which expresses the channel wise second-order statistical information of the feature, thus achieving independence from scene structure while effectively modeling inter-channel relationships. For feature \( \textbf{X} \in \mathbb{R}^{a \times b} \), the Gram matrix is defined as:

\begin{equation}
\text{Gram}(\textbf{X}) = \frac{1}{a \times b} \textbf{X} \cdot \textbf{X}^T,
\end{equation}
where $a$ denotes the feature dimension and $b$ denotes the number of feature vectors.

Based on this theory, we flatten the processed style feature $\textbf{S}_\text{1}$ and the style feature $\textbf{S}_\text{gt}$ of the ground truth in the spatial dimension, and then calculate the similarity of their Gram matrices to supervise the style features to accurately characterize the degradation patterns. The style decoupling loss is as follows.
\begin{equation}
\mathcal{L}_{\text{style}} = (\text{Gram}(\textbf{S}_\text{1}) - \text{Gram}(\textbf{S}_\text{gt}))^2.
\end{equation}

Combined with the reconstrcution losses, The total training objective in representation decoupling is:
\begin{equation}
\begin{split}
\mathcal{L}_{\text{rep-dec}} & = \mathcal{L}_{\text{recon}}({I}_0, \hat{I_0}) +  \mathcal{L}_{\text{recon}}({I}_\text{gt}, \hat{I_\text{gt}}) \\
& +  \mathcal{L}_{\text{recon}}({I}_\text{gt}, \hat{I_1}) + \lambda_{\text{style}} \mathcal{L}_{\text{style}},
\end{split}
\end{equation}
where $\lambda_{\text{style}}$ denotes the weight of style losss, \({I}_0\) is the input degraded image, \({I}_\text{gt}\) is the corresponding target image, \(\hat{I_0}\) and \(\hat{I_\text{gt}}\) denotes the corresponding reconstruction result, and $\hat{I_1}=\text{D}(\textbf{C}_1)$ denotes the enhancement result decoded from $\textbf{C}_1$.

Through network function decomposition and style transfer supervision, the representation decoupling process utilizes a style feature to capture the degradation pattern and models its evolution process through the SREU. This provides a principled condition for subsequent adaptive trajectory modulation.

\subsection{Adaptive trajectory modulation}\label{subsec:mar}
After modeling the degradation patterns and their evolution process through representation decoupling, we perform adaptive enhancement trajectory adjustment. We found that the enhancement process is not a uniform mapping, but a unique trajectory that varies according to the diverse image degradation situations. Specifically, for mildly degraded images, a standard enhancement may lead to over-processing, implying that the ideal restoration resides on an interpolative point between the input and the model's nominal output. Conversely, for severely degraded samples, an uniform mapping often proves to be insufficient, whereas recursive application of the same enhancement logic can progressively drive the images toward a high-fidelity state. Based on this intuition, we conceptualize the enhancement process as a continuous state evolution conditioned on the degradation patterns.

To realize this evolution, the SREU functions as a shared recursive enhancement unit that iteratively refines the features, generating a sequence of candidate states along a potential enhancement trajectory. In order to obtain the most ideal enhancement results from the potential enhancement trajectory, we introduce Ada-Route module that conditioned on the modeled degradation patterns to perform a weighted fusion of these candidate states.  
To train such a router, we need additional supervision signals to guide it in assigning higher weights to states that are closer to the ideal enhancement. This supervision signal is obtained by self-evaluating during the training process.


\subsubsection{Recursive enhancement trajectory and Ada-Route module}\label{sec:Adaptive routing}
Inspired by the observation that a secondary enhancement can further refine insufficient restoration, we employ a recursive strategy to construct enhancement trajectories. Building upon the architecture introduced in Sect. \ref{subsec:mst}, which consists of E, D, and SREU, we define a basic UIE network as $\mathcal{N}(\cdot)=\text{D}(\text{SREU}(\text{E}(\cdot)))$. As this network learns an uniform mapping during representation decoupling, it also faces the challenge of sub-optimal enhancement. Formally, our observation can be expressed as $\mathcal{Q}(\mathcal{N}(\mathcal{N}(I_{0}))) > \mathcal{Q}(\mathcal{N}(I_{0}))$ in specific scenarios, where $\mathcal{Q}$ denotes a quality evaluator. Given that robust image representations are already obtained via reconstruction trainings, we simplify the recursive application of the entire network into a recursive invocation of the SREU:

\begin{equation}
\{\textbf{C}_k,\textbf{S}_k\} =
\begin{cases}
\text{E}(I_0), & k = 0, \\
\text{SREU}(\textbf{C}_{k-1},\textbf{S}_{k-1}), & k = 1, \dots, K,
\end{cases}
\end{equation}
where $k \in \{0, 1, \dots, K\}$ is the number of enhancement iterations and $K$ is the maximum iteration times. In practice, we find that trajectories beyond two iterations yield diminishing returns, therefore, we set $K = 2$ as our default implementation.
From this, we constructed a recursive state transition trajectory in the image representation space and degradation pattern space driven by SREU.

\begin{figure}
	\centering
	\includegraphics[width=1.0\columnwidth]{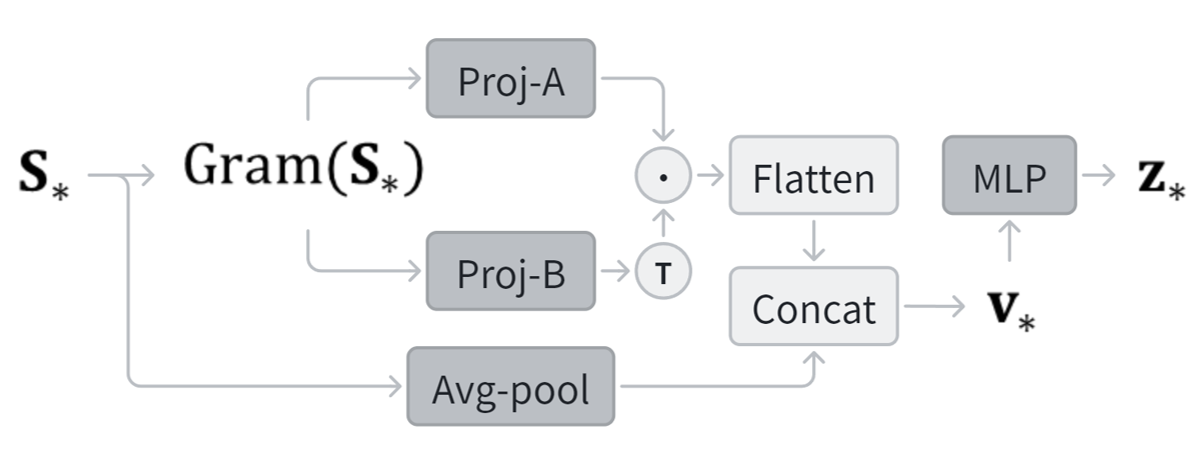}
	\caption{Design of Ada-Route module. T denotes transpose, Proj-A and Proj-B denotes linear projection layer aiming to compress the information from Gram matrix.}
	\label{FIG:5}
\end{figure}

As the SREU iteratively evolves the features from a degraded state to a clean state, the modeled degradation patterns serve as adaptive routing signals that represent the enhancement situations in each state. To utilize these signals, we introduce the Ada-Route as illustrated in Fig. \ref{FIG:5}. The Ada-Route is designed to perceive the current state of restoration by analyzing the degradation-aware style feature $\textbf{S}_*$ of the candidate states. Specifically, for each candidate state, the module takes the corresponding style feature as input. 
Since we model the degradation pattern in the representation decoupling by supervising the changes of the Gram matrix, we also use the Gram matrix to determine the degradation condition during routing. In addition, we also use the global features obtained by global average pooling to assist in routing.
Since the Gram matrix is too large when directly flattened for an MLP, we used two projection layers to compress the information of the Gram matrix. The compressed Gram matrix is then flattened, concatenated with the global vector, and fed into an MLP to produce the corresponding logit $\textbf{z}_*$:
\begin{equation}
\mathbf{z}_k = \text{Ada-Route}(\textbf{S}_{k}), k = 0, 1,\dots, K.
\end{equation}

The logits obtained  are concatenated and passed through a softmax layer to yield a probability distribution over candidate states:
\begin{equation}
\mathbf{w} = \text{softmax}\big([\mathbf{z}_0, \mathbf{z}_1 ... \mathbf{z}_K]\big) \in \mathbb{R}^3.
\end{equation}

Then, the weights obtained $\mathbf{w} = [w_0, w_1, w_2]$ are used to perform a weighted fusion of the image representations of the candidate states, obtaining the final adaptively adjusted image representation $\textbf{C}_{\text{w}}$:

\begin{equation}
\textbf{C}_{\text{w}} = \sum_{k=0}^{K} w_k \, \textbf{C}_{k}.
\end{equation}

Finally, we decode the adaptive adjusted image representation $\textbf{C}_{\text{w}}$ to get the final enhancement result:
\begin{equation}
\hat{I_{\text{w}}}=\text{D}(\textbf{C}_{\text{w}}).
\end{equation}

\subsubsection{Pseudo-labeling and training strategy}\label{sec:pseudo_label}
To enable the training of the Ada-Route module, we need to obtain additional signals that can indicate the potential optimal choice. Based on our observation, we use recursive enhancement to produce robust and intuitive labels.
We formulate pseudo-label generation as a selection problem over the number of enhancement iterations $k \in \{0, 1, \dots, K\}$. During the late training phase, we perform self-evaluation within each training iteration. Starting from the input image $I_{0}$, we use the basic UIE network $\mathcal{N}$ defined in Sect. \ref{sec:Adaptive routing} to generate a cascade of enhanced outputs:
\begin{equation}
I_{k} =
\begin{cases}
I_{0}, & k = 0, \\
\mathcal{N}\big(I_{k-1}\big), & k = 1, \dots, K.
\end{cases}
\end{equation}

The quality evaluator $\mathcal{Q}(\cdot)$ assigns a scalar score to each candidate: $e_k = \mathcal{Q}\big(I_{k}\big)$. In our implementation, the $\mathcal{Q}(\cdot)$ is the PSNR metric. The optimal number of enhancement iterations is then selected as
\begin{equation}
\bar{k} = \argmax_{k \in \{0, \dots, K\}} e_k.
\end{equation}

This label serves as the pseudo-label and provides supervision for adaptive routing. 
To supervise this routing decision, we minimize the cross-entropy loss between the predicted logits $\mathbf{z}_*$ and the pseudo-label $\bar{k} \in \{0,1 ...K\}$:
\begin{equation}
\mathcal{L}_{\text{route}} = \mathrm{CE}\big( [\mathbf{z}_0, \mathbf{z}_1 ... \mathbf{z}_K],\, \bar{k} \big),
\end{equation}
where $\mathrm{CE}(\cdot, \cdot)$ denotes the standard cross-entropy loss that takes raw logits and a scalar class index as input.
This loss explicitly supervises the router to modulate enhancement trajectory by aligning routing decisions with the possible optimal candidate results derived from pseudo-labeling.

Given the potential for distortion due to iterative calls of the SREU, we introduce a targeted reconstruction loss adjustment mechanism for samples with varying enhancement trajectories. Specifically, we calculate the reconstruction loss between the final weighted fusion result and the result of the possibly optimal state with the ground truth. We consider the pseudo-label $\bar{k}$ as the possibly optimal state within the enhancement trajectory, and compute the reconstruction loss by aligning the image representation at the $\bar{k}$-th iteration with the ground truth:
\begin{equation}
\mathcal{L}_{\text{k-recon}} = 
\begin{cases}
0, & \bar{k} = 0, \\
\mathcal{L}_{\text{recon}}\big(I_{\text{gt}}, \text{D}(\textbf{C}_{\bar{k}})\big), & \bar{k} = 1, \dots, K.
\end{cases}
\end{equation}

Similarity, the reconstruction loss for the final fused feature is computed as:
\begin{equation}
\mathcal{L}_{\text{w-recon}} = \mathcal{L}_{\text{recon}}\big(I_{\text{gt}}, \hat{I_\text{w}}\big).
\end{equation}

Combining with the router loss and style decoupling loss, the final training objective in adaptive trajectory modulation is:
\begin{equation}
\begin{split}
\mathcal{L}_{\text{ada-mod}} & = \lambda_{\text{rep-dec}}\mathcal{L}_{\text{rep-dec}} +  \lambda_{\text{w-recon}}\mathcal{L}_{\text{w-recon}}\\
& +\lambda_{\text{route}}\mathcal{L}_{\text{route}} +  \lambda_{\text{k-recon}}\mathcal{L}_{\text{k-recon}}.
\end{split}
\end{equation}
where $\lambda_\text{rep-dec}$, $\lambda_\text{w-recon}$, $\lambda_\text{route}$ and $\lambda_\text{k-recon}$ are the weights corresponding to each loss.

By jointly optimizing the losses above, our framework learns an adaptive enhancement trajectory routing mechanism conditioned on the learned degradation pattern from representation decoupling, enabling the model to perform optimal enhancement for each input.

\begin{table*}
\caption{Quantitative comparison on UIEB and LSUI dataset. Higher values are better. Best and second-best results are \textbf{bolded} and \underline{underlined}, respectively.}
\label{tab:main_results}
\begin{tabular*}{\tblwidth}{@{}LCCCCCCCCCC@{}}
\toprule
\multirow{2}{*}{Method} & \multicolumn{5}{c}{UIEB} & \multicolumn{5}{c}{LSUI} \\
\cmidrule(r){2-6} \cmidrule(l){7-11}
& PSNR$\uparrow$ & SSIM$\uparrow$ & UIQM$\uparrow$ & UCIQE$\uparrow$ & MUSIQ$\uparrow$ 
& PSNR$\uparrow$ & SSIM$\uparrow$ & UIQM$\uparrow$ & UCIQE$\uparrow$ & MUSIQ$\uparrow$ \\
\midrule
UDCP & 11.0736 & 0.5153 & 3.6838 & 0.6149 & 59.3019 & 13.1057 & 0.5623 & 4.1156 & \underline{0.5982} & 47.6424 \\
Ucolor & 19.9437 & 0.7642 & 4.1840 & 0.5879 & 55.1289 & 20.1013 & 0.7003 & 3.9241 & 0.5572 & 35.8873 \\
UWCNN  & 20.0796 & 0.8933 & 4.3106 & 0.5808 & 60.5273 & 23.4756 & 0.8944 & 4.3802 & 0.5716 & 53.4860 \\
U-shape & 22.6897 & 0.8391 & 4.3122 & 0.6205 & 56.5789 & 26.9908 & 0.8765 & 4.3781 & 0.5905 & 51.2496 \\
DPF-Net & 23.0872 & 0.8518 & 4.2407 & 0.6056 & \textbf{63.7496} & 21.8208 & 0.8345 & 4.3645 & \textbf{0.6015} & 49.9639 \\
DW-Net & 23.3311 & 0.9200 & 4.3067 & 0.6047 & 60.7413 & 27.0938 & 0.9123 & 4.3575 & 0.5891 & 52.6896 \\
GUPDM & 23.9009 & 0.9277 & \underline{4.3378} & 0.6195 & 60.0297 & 28.9015 & 0.9289 & 4.3943 & 0.5931 & 54.1847 \\
HCLR-Net & 24.0709 & 0.9264 & \textbf{4.3657} & \textbf{0.6274} & 61.6649 & 29.0262 & 0.9120 & \textbf{4.4073} & 0.5976 & \textbf{57.0729} \\
Semi-UIR (\textit{.semi}) & 24.8590 & \underline{0.9334} & 4.3337 & 0.6212 & 61.6020 & 25.8441 & 0.8945 & \underline{4.4007} & 0.5971 & \underline{55.7371} \\
Semi-UIR (\textit{.sup}) & \underline{25.1202} & 0.9325 & 4.2942 & 0.6223 & 61.0746 & \underline{29.2363} & \underline{0.9273} & 4.3717 & 0.5946 & 55.3701 \\
\midrule
\textbf{SDAR-Net} & \textbf{25.7249} & \textbf{0.9372} & 4.2958 & \underline{0.6244} & \underline{61.8258} & \textbf{30.3338} & \textbf{0.9274} & 4.3884 & 0.5950 & 55.5414 \\
\bottomrule
\end{tabular*}
\end{table*}

\section{Experiments}

\subsection{Experiments setup}\label{sec:Datasets and evaluations}
\textbf{Datasets.} We evaluate our method on two widely used real-world underwater image enhancement datasets:  
UIEB~\citep{UIEB}, which contains 890 paired underwater images with various degradation situations, and LSUI~\citep{U-shape}, a large-scale paired dataset comprising 4279 underwater images. The images in these datasets do not have a uniform resolution, following previous studies, we reshaped all images to a resolution of $256\times256$ for all training and testing.

\textbf{Evaluation metrics.} Following standard practice, we report both full-reference and no-reference metrics to comprehensively assess perceptual and structural quality. Full-reference metrics include PSNR and SSIM \citep{wang2004image}, which can effectively reflect the reconstruction accuracy of supervised training methods. For no-reference metrics, we employ three widely adopted no-reference metrics in underwater image enhancement studies:  
UIQM \citep{panetta2015uiqm}, UCIQE \citep{yang2015uciqe}, and MUSIQ \citep{ke2021musiq}. UIQM \citep{panetta2015uiqm} and UCIQE \citep{yang2015uciqe} are evaluation metrics based on statistical information of the image, primarily measuring the richness and naturalness of color in underwater images. MUSIQ \citep{ke2021musiq} is an image quality scoring method based on deep learning. Higher values indicate better quality for all metrics.

\textbf{Comparison Methods.} For the comparative method, we select a representative suite of models spanning various paradigms: the traditional UDCP \citep{drews2016underwater} serves as a baseline, while deep learning methods include CNNs such as UWCNN \citep{UWCNN}) and Ucolor \citep{Ucolor}, GAN-based Transformer hybrids like U-shape \citep{U-shape}, and physical-prior-informed model such as GUPDM \citep{GUPDM} and DPF-Net \citep{mei2025dpf}. We also compare against recent CNN and U-Net hybrid architectures specifically optimized for UIE, such as DW-Net \citep{DW-net}, HCLR-Net \citep{HCLR-Net}, and the semi-supervised Semi-UIR \citep{Semi-UIR}. 
\begin{figure*}
	\centering
	\includegraphics[width=1\textwidth]{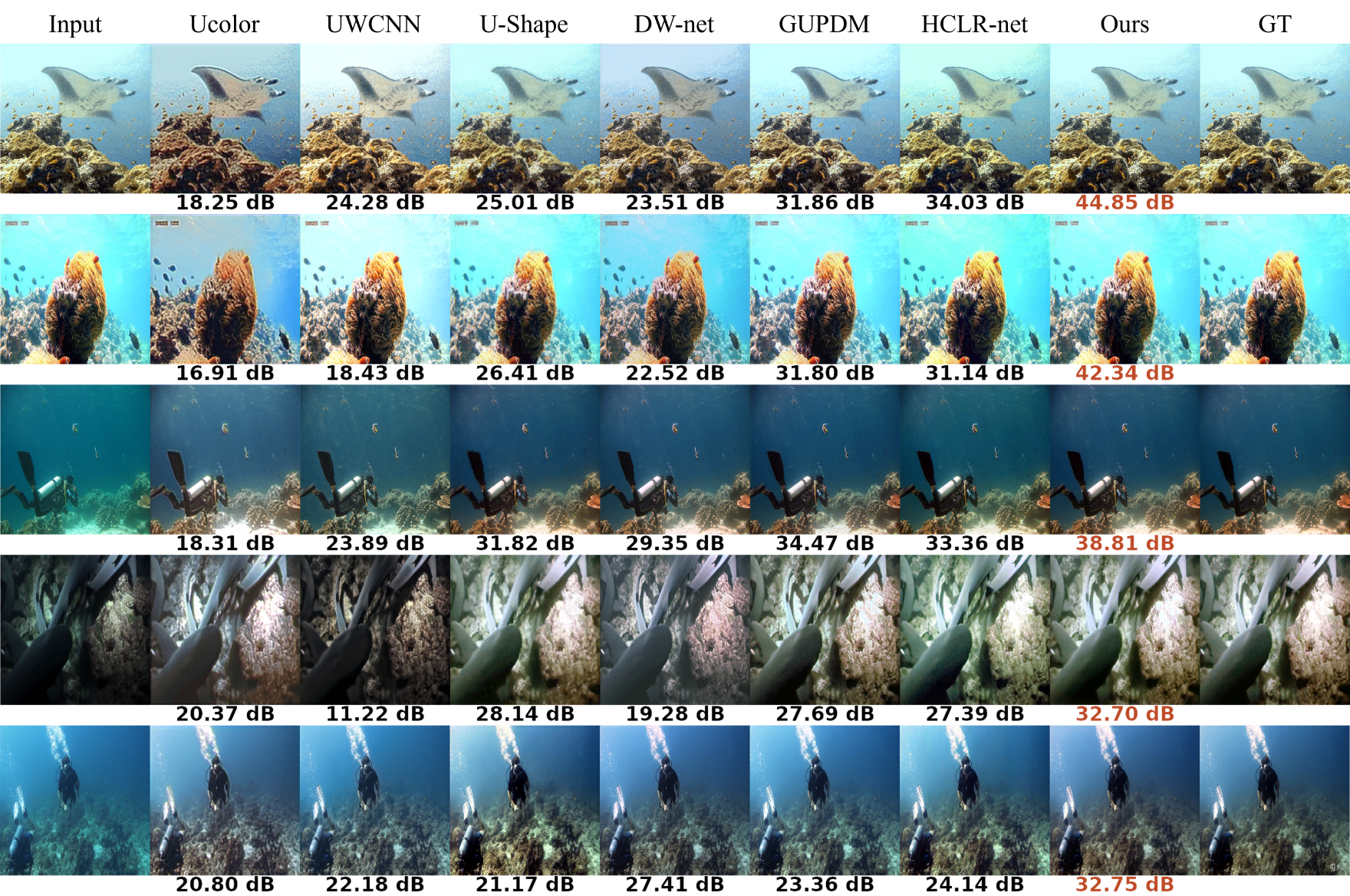}
	\caption{Qualitative comparison on UIEB dataset. We mark the output image with the corresponding PSNR.}
	\label{FIG:6}
\end{figure*}
\subsection{Main Performance}\label{sec:Results on real-world datasets}

Quantitative comparisons with state-of-the-art methods are summarized in Table~\ref{tab:main_results}. 
Our SDAR-Net achieves the best performance on full-reference metrics (PSNR and SSIM) across both UIEB~\citep{UIEB} and LSUI~\citep{U-shape}, indicating superior reconstruction fidelity with respect to reference images. 
Since no-reference metrics assess image quality from certain perspectives, such as pixel statistics and deep learning-based scores, it is difficult for them to evaluate a model's preservation of scene structure. However, maintaining the scene structure is crucial for downstream tasks, and its importance is validated in Sect. \ref{subsec:Results on Down Stream Tasks}. Therefore, we primarily compare performance based on reference metrics. Nevertheless, we also achieve results comparable to existing methods on no-reference metrics, indicating that our method considers visual quality while effectively preserving scene structure in diverse underwater degradation scenarios.
Qualitative results in Fig.~\ref{FIG:6} further confirm that SDAR-Net excels in handling different degradation situations, particularly in edge-cases where our adaptive trajectory routing adaptively invokes different enhancement states to mitigate various degradations. In the first two rows of Fig.~\ref{FIG:6}, mildly degraded images are well reconstructed, avoiding color distortion and noise amplification caused by excessive processing, thereby significantly improving accuracy. The last three lines show that we can maintain robust recovery across various degradation scenarios and levels of degradation.
\begin{figure*}
\centering
\includegraphics[width=1\textwidth]{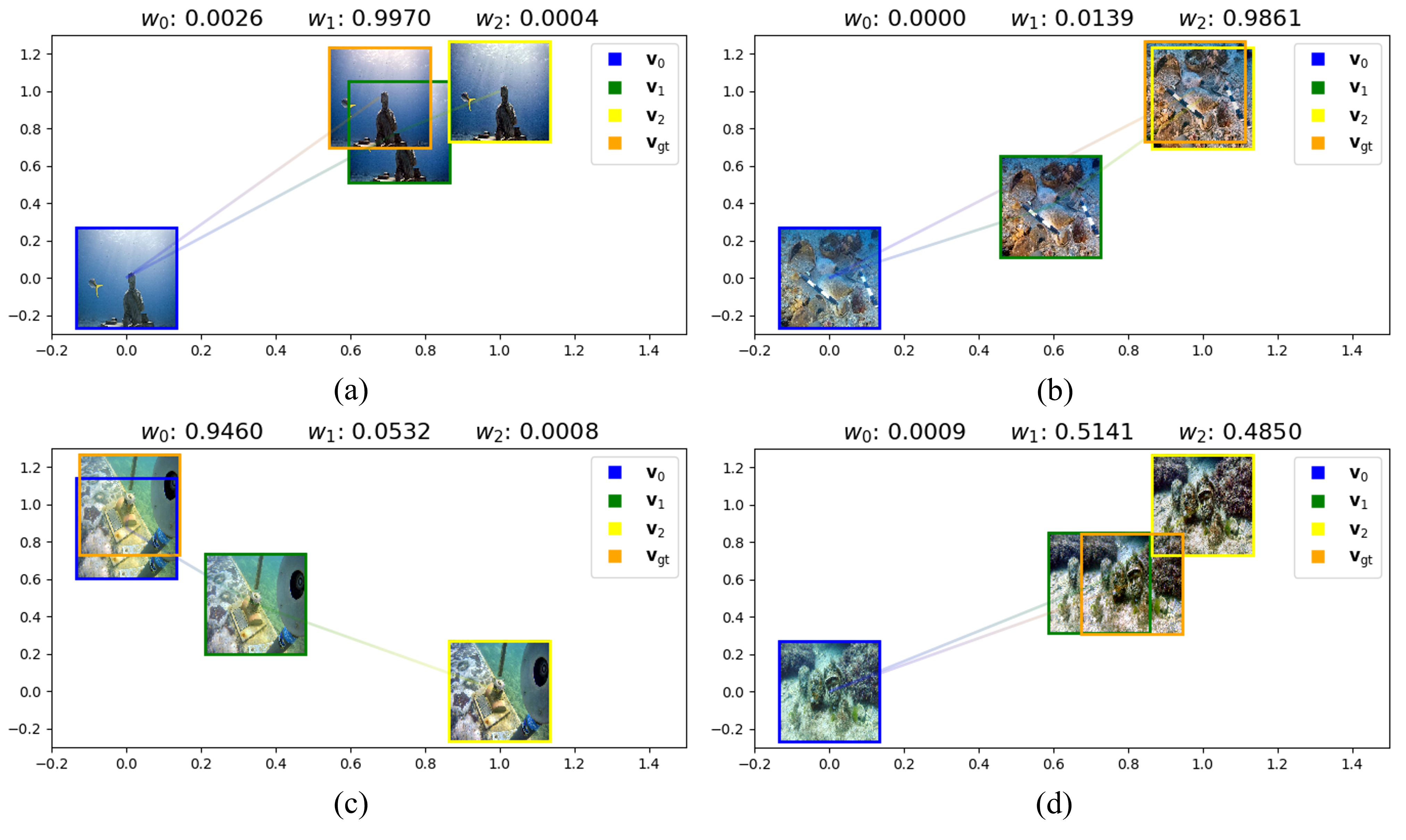}
\caption{
Enhancement trajectory visualize with routing condition vectors. (a) represents the common cases that can be covered by state 1. (b) represents cases that need a deeper enhancement trajectory. (c) represents cases that only requires a small degree of enhancement. (d) represents that the optimal trajectory is between two candidate states.
}
\label{fig:direction}
\end{figure*}
\subsection{Ablation studies}\label{subsec:ablation}
\begin{table}
\centering
\caption{Ablation study on UIEB. We evaluate all combinations of the representation decoupling (Decouple) and the adaptive trajectory modulation (Ada-Route).}
\label{tab:ablation_components}
\begin{tabular*}{\tblwidth}{@{}CCCC@{}}
\toprule
w/o Decouple & w/o Ada-Route & PSNR$\uparrow$ & MUSIQ$\uparrow$ \\
\midrule
\texttimes & \texttimes & 25.0219 & 61.8405 \\
\checkmark & \texttimes & 25.3101 & 61.7947  \\
\texttimes & \checkmark & 25.0065 & \textbf{62.1304} \\
\checkmark & \checkmark & \textbf{25.7249} & 61.8258 \\
\bottomrule
\end{tabular*}
\end{table}
\subsubsection{Component-wise ablation}\label{subsec:Component-wise Ablation}
To validate the design choices in SDAR-Net, we conduct ablation studies on UIEB~\citep{UIEB} dataset. The results are shown in Table~\ref{tab:ablation_components}. 
We first examine the contribution of each major component: the representation decoupling (denoted ``Decouple'') the adaptive trajectory modulation (``Ada-Route''). 

The ablation of the representation decoupling module is specifically reflected in the removal of the style decoupling loss. Without this guidance, the learning of the enhancement module SREU is no different from a conventional unified mapping, making it difficult to explicitly model the core changes of degradation. The ablation of adaptive trajectory modulation is reflected in not performing iterative enhancement trajectory construction and removing the Ada-Route module, making the model incapable of explicitly constructing different enhancement processes. It is worth noting that, in the absence of representation decoupling, the adaptive trajectory modulation is difficult to be effective. This indicates that the lack of explicit degradation pattern modeling makes it difficult for the routing module to determine whether the enhancement effect at the current state meets expectations or whether further enhancement is needed. In summary, removing either component leads to significant performance drops, confirming that both explicit representation decoupling for modeling degradation patterns and adaptive trajectory modulation are essential.

\subsubsection{Number of enhancement iterations}\label{subsec:Number of Enhancement Pathways}

We investigate the impact of the maximum  number of enhancement iterations $K$ on the UIEB dataset. We conducted separate training using different $K$. As shown in Table~\ref{tab:ablation_pathways}, the configuration with \(K = 2\) achieves the best performance across all metrics. Although increasing \(K\) from 3 to 4 yields an improvement over \(K = 3\), its results remain consistently inferior to those of \(K = 2\).

This raises the question of whether an increase in \(K\) could eventually surpass the performance of \(K = 2\). To address this, we analyze the average routing weights assigned to each state. Crucially, for \(K = 4\), the average weight of the deepest trajectory \(w_4\) is nearly zero, indicating that the model effectively discards this additional state. This suggests that the enhancement process has already converged by \(k=2\), and further refinements do not offer a meaningful contribution. Consequently, adding more iterates beyond \(K = 2\) is unlikely to improve performance while unnecessarily increasing computational cost.

\begin{table}
\centering
\caption{Ablation on the number of enhancement iterations.}
\label{tab:ablation_pathways}
\begin{tabular*}{\tblwidth}{@{}CCCCCCCC@{}}
\toprule
$K$ & PSNR$\uparrow$ 
& MUSIQ$\uparrow$ 
& $w_0$ & $w_1$ & $w_2$ & $w_3$ & $w_4$\\
\midrule
1 & 25.4759 
& 62.0685 
& 0.0449 & 0.9551 & - & - & - \\
2 & \textbf{25.7249} 
& \textbf{61.8258} 
& 0.0311 & 0.8201 & 0.1488 & - & - \\
3 & 25.4688 
& 61.3879 
& 0.0414 & 0.8461 & 0.1097 & 0.0028 & - \\
4 & 25.6128
& 61.5108 
& 0.0487 &0.8874 & 0.0504 & 0.0134 & 0.0001 \\
\bottomrule
\end{tabular*}
\end{table}

\begin{figure*}
	\centering
	\includegraphics[width=0.76\textwidth]{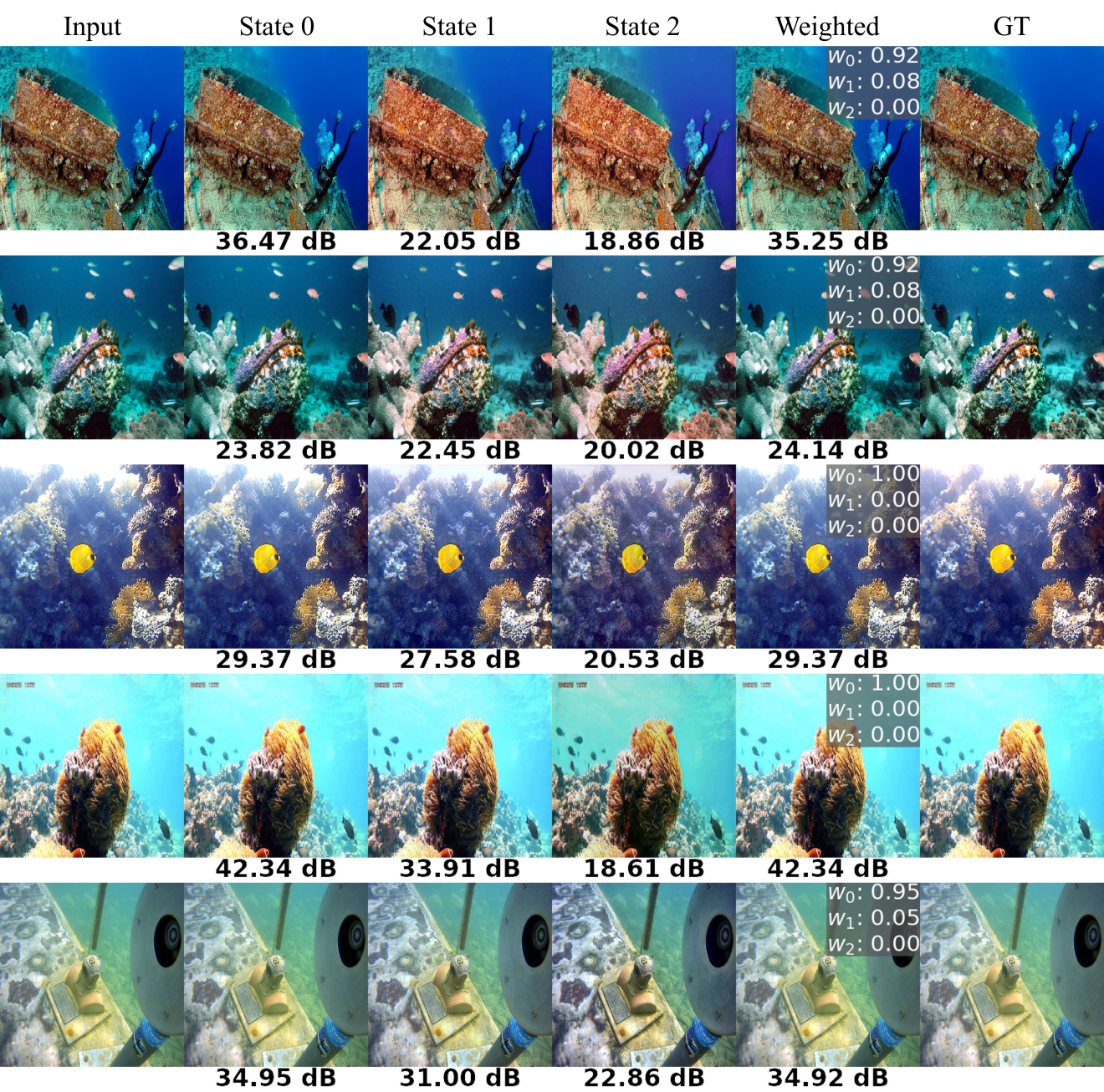}
	\caption{Routing effect on mild degradation cases. We mark the output image with the corresponding PSNR and weighted results with its routing weights.}
	\label{FIG:7a}
\end{figure*}

\begin{figure*}
	\centering
	\includegraphics[width=0.76\textwidth]{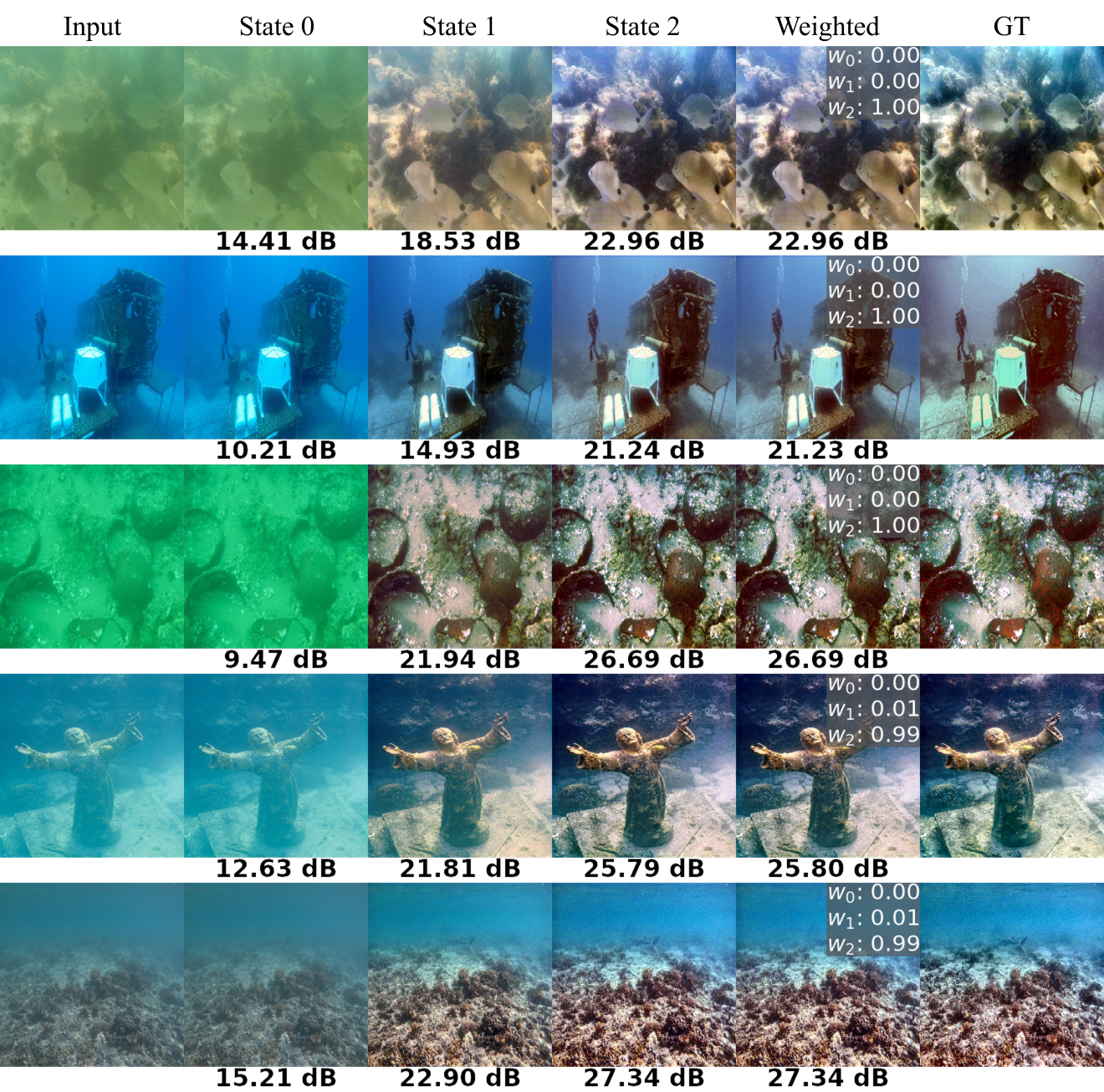}
	\caption{Routing effect on severe degradation cases. We mark the output image with the corresponding PSNR and weighted results with its routing weights.}
	\label{FIG:7b}
\end{figure*}

\subsection{Analysis of degradation pattern modeling}\label{subsec:direction_analysis}

To gain deeper insights into how our framework perceives and modulates the restoration process, we visualize the degradation-aware style features in a latent decision manifold. Specifically, for each state's style feature $\textbf{S} \in \{\textbf{S}_0, \textbf{S}_\text{1}, \textbf{S}_\text{2}, \textbf{S}_\text{gt}\}$, we map its routing vector $\textbf{v} \in \{\textbf{v}_0, \textbf{v}_1, \textbf{v}_2, \textbf{v}_\text{gt}\}$ into a 2D coordinate $(x, y)$ using PCA. 

As illustrated in Fig. \ref{fig:direction}, a key observation is the strong directional alignment between the sequence $\{\textbf{v}_0, \textbf{v}_\text{1}, \textbf{v}_\text{2}\}$ and the ideal restoration vector $\{\textbf{v}_0, \textbf{v}_\text{gt}\}$. This consistent trajectory suggests that our degradation-aware style modeling successfully captures the intrinsic stylistic shift from a degraded state to a clear one. 
Complementing this directional consistency, the Ada-Route mechanism effectively senses restoration progress by analyzing the relative proximity of these candidate states to the target. We observe that the weights allocated to candidate states closer to $\textbf{v}_\text{gt}$ receive significantly higher priority, indicating that the router's ability to accurately prioritize the most refined features is derived from the well modeling of degradation patterns. This behavior confirms the importance of obtaining the critical characteristics of degradation and reflects them in the style feature space. 

\subsection{Visualization of routing effect}
We demonstrate the effectiveness of the routing mechanism by visualizing the weights corresponding to the intermediate states in edge-cases.
In Fig.~\ref{FIG:7a} and Fig.~\ref{FIG:7b}, we decode the image representations corresponding to different candidate states and compare them with the effect of weighted fusion. Among them, ``State 0'' corresponds to the reconstruction of the input image, ``State 1'' is to the result under the usual uniform mapping process, and ``State 2'' represents the effect after secondary processing. In the ``Weighted'' column, we display the routing weights $(w_0, w_1, w_2)$ for the corresponding states ($k=0,1,2$). It can be observed that our designed routing mechanism accurately handles these edge-cases by selectively invoking different enhancement states. Specifically, our design can effectively identify the degradation patterns in different underwater scenes, thereby distinguishing the current degradation situation and giving higher weight to the appropriate enhanced state.

\subsection{The necessity of adaptive routing mechanism}
We further explore why an explicit modulate mechanism is needed, while implicit processing or simply reusing models struggles to achieve adaptive effects.
We first train a model $\mathcal{N}'(\cdot)=\text{D}(\text{SREU}(\text{SREU}(\text{E}(\cdot))))$ that recursively calls the SREU module twice to test whether the SREU module itself can adapt to different underwater degradations without explicit adjustment, which is denoted as Internal cascade in Table \ref{tab:state}. 
Then, the External cascading in Table \ref{tab:state} refers to feed the enhanced results of the model back into the model to obtain a second-level enhanced result. We use the basic model $\mathcal{N}$  described in
Sect. \ref{sec:pseudo_label} and apply it twice to test whether the overall model can adapt to different inputs.
Models used in both external and internal cascade are trained with representation decoupling but without adaptive trajectory modulation. Additionally, we decode the results from the candidate state representations in the full model for reference, denoted as State 0, 1, and 2 in the strategies column.

As shown in Table \ref{tab:state}, our complete adaptive routing method is superior to all other solutions. The external cascade scheme greatly reduces the performance, because the model itself lacks adaptability, resulting in distortion when processing images that are already clear. The internal cascade scheme also does not have a significant improvement compared to the ``State 1'' mapping, indicating that it is essentially a unified mapping, with insufficient handling of edge cases.
Therefore, our explicitly designed adaptive approach that integrates various enhanced states is the most effective scheme.

\begin{table}
\centering
\caption{Analysis of routing effectiveness. 
}
\label{tab:state}
\begin{tabular*}{\tblwidth}{@{}CCC@{}}
\toprule
Strategies & PSNR$\uparrow$ & MUSIQ$\uparrow$ \\
\midrule
State 0 & 17.3298 & 59.6402 \\
State 1 & 25.4238 & 61.8111  \\
State 2 & 22.3304 & 61.5854 \\
\midrule
External cascade & 22.0784 & 61.7479 \\
Internal cascade & 25.4681 & 61.7842 \\
\midrule
Weighted & \textbf{25.7249} & \textbf{61.8258} \\
\bottomrule
\end{tabular*}
\end{table}

\subsection{Results on down stream tasks}\label{subsec:Results on Down Stream Tasks}
To further validate the effectiveness of the proposed method, we evaluate its impact on underwater semantic segmentation and compare it against several state-of-the-art underwater image enhancement approaches.

We conduct experiments on the SUIM dataset \citep{islam2020semantic}, which contains 1635 images annotated with eight semantic categories: background waterbody (BW), human divers (HD), plants / sea-grass (PF), wrecks / ruins (WR), robots / instruments (RO), reefs / invertebrates (RI), fish / vertebrates (FV), and sand / sea-floor (SR). Following standard practice, we adopt a pretrained U-Net architecture \citep{ronneberger2015u} as the segmentation backbone. All training and test images are first enhanced using the compared UIE methods, after which the segmentation network is fine-tuned and evaluated on the enhanced data.

As shown in Table~\ref{tab:DS_results}, our method yields the highest average segmentation performance. By adaptively correcting color distortion and improving contrast in degraded underwater scenes, SDAR-Net steadily enhances the visual discriminability of different object classes in diverse underwater scenes, thereby facilitating more accurate semantic predictions. 

\begin{table*}
\caption{Quantitative comparison of effectiveness on underwater semantic segmentation. The metric is mIoU, higher values are better. Best and second-best results are \textbf{bolded} and \underline{underlined}, respectively.}
\label{tab:DS_results}
\begin{tabular*}{\tblwidth}{@{}LCCCCCCCCC@{}}
\toprule
Method & Avg & BW & HD & PF & WR & RO & RI & FV & SR \\

\midrule
Baseline & 0.6046 & 0.8457 & 0.6686 & \underline{0.2107} & 0.5672 & 0.6092 & 0.6049 & 0.6581 & \underline{0.6723} \\
Ucolor  & 0.5780 & 0.8401 & 0.6657 & 0.0196 & 0.6205 & 0.6107 & 0.6121 & 0.6428 & 0.6122 \\
U-shape  & 0.5831 & \textbf{0.8584} & 0.7130 & 0.0578 & 0.6275 & 0.4654 & 0.6017 & 0.6984 & 0.6428 \\
UDCP & 0.5903 & 0.8512 & 0.6800 & 0.0716 & 0.5686 & 0.6384 & 0.5876 & 0.7066 & 0.6182 \\
UWCNN  & 0.6051 & 0.8540 & 0.7023 & 0.2073 & 0.6142 & 0.5646 & 0.6072 & 0.6458 & 0.6457 \\
HCLR-Net & 0.6125 & 0.8492 & 0.6669 & \textbf{0.2344} & 0.6061 & 0.6137 & 0.6309 & 0.6966 & 0.6017 \\
DW-Net   & 0.6195 & 0.8541 & 0.6688 & 0.1544 & \underline{0.6694} & 0.6264 & 0.6288 & 0.6884 & 0.6656 \\
Semi-UIR   & 0.6198 & 0.8354 & \underline{0.7152} & 0.1762 & 0.6459 & \underline{0.6507} & \underline{0.6322} & \underline{0.7110} & 0.5920 \\
GUPDM   & \underline{0.6267} & \underline{0.8572} & 0.6504 & 0.1972 & \textbf{0.6727} & 0.6312 & \textbf{0.6557} & 0.6693 & \textbf{0.6803} \\
\midrule
\textbf{SDAR-Net} & \textbf{0.6358} & 0.8569 & \textbf{0.7303} & 0.1822 & 0.6559 & \textbf{0.7052} & 0.5758 & \textbf{0.7337} & 0.6464 \\
\bottomrule
\end{tabular*}
\end{table*}

\section{Conclusion} 
In this paper, we have presented SDAR-Net, a novel adaptive enhancement framework that departs from conventional uniform paradigms. Based on prior knowledge that underwater degradation primarily affects appearance rather than scene structure, we decoupled the degradation-aware style embeddings from image representations through a specialized training framework. Then we introduced an adaptive routing mechanism, which adaptively modulates the enhancement process via soft-weight prediction to satisfy the specific restoration demands of each image. Extensive evaluations of real-world benchmarks demonstrate that SDAR-Net achieves state-of-the-art performance, notably 
achieving a PSNR of 25.72 dB on real-world benchmark 
and showing significant utility in downstream vision tasks such as semantic segmentation.

\printcredits

\section*{Acknowledgements}
This work was supported by the Wuhan Natural Science Foundation Project (2025041001010363).


\bibliographystyle{cas-model2-names}

\bibliography{cas-refs}



\end{document}